%% file: 2314.tex
\documentclass[runningheads]{llncs}
\usepackage{graphicx}
\usepackage{amsmath,amssymb} 
\usepackage{color}
\usepackage{bbm}
\usepackage{booktabs}
\usepackage{cleveref}
\usepackage{algorithm,algorithmic}
\usepackage{symbols}
\usepackage{multirow}

\begin{document}
\pagestyle{headings}
\mainmatter

\title{Straight to the Facts: Learning Knowledge Base Retrieval for Factual Visual Question Answering} 

\titlerunning{Learning Knowledge Base Retrieval for Factual Visual Question Answering}

\authorrunning{Medhini Narasimhan and Alexander G. Schwing}

\author{Medhini Narasimhan, Alexander G. Schwing}


\institute{University of Illinois Urbana-Champaign\\
	\email{ \{medhini2,aschwing\}@illinois.edu}
}
\maketitle

\begin{abstract}
Question answering is an important task for autonomous agents and virtual assistants alike and was shown to support the disabled in efficiently navigating an overwhelming environment. Many existing methods focus on observation-based questions, ignoring our ability to seamlessly combine observed content with general knowledge. To understand interactions with a knowledge base, a dataset has been introduced recently and keyword matching techniques were shown to yield compelling results despite being vulnerable to misconceptions due to synonyms and homographs. To address this issue, we develop a learning-based approach which goes straight to the facts via a learned embedding space. We demonstrate state-of-the-art results on the challenging recently introduced fact-based visual question answering dataset, outperforming competing methods by more than $5\%$.

\end{abstract}

\input{introduction}

\input{related}
\input{method}

\input{evaluation}
\input{conclusion}

\bibliographystyle{splncs}
\bibliography{ref}
\newpage
\end{document}

%% file: introduction.tex
\section{Introduction}
When answering questions given a context, such as an image, we seamlessly combine the observed content with general knowledge. 
For autonomous agents and virtual assistants which 
naturally participate in our day to day endeavors, where answering of questions based on context and general knowledge is most natural,  algorithms which leverage both observed content and general knowledge are extremely useful. 

To address this challenge, in recent years, a significant amount of research has been devoted to question answering in general and Visual Question Answering (VQA) in particular. Specifically, the classical VQA tasks require an algorithm to answer a given question based on the additionally provided context, given in the form of an image. For instance, significant progress in VQA was achieved by introducing a variety of VQA datasets with strong baselines \cite{VisGen,RenNIPS2015,ZhuCVPR2016,visturing,JohnsonCVPR2017Clevr,JabriARXIV2016,YuARXIV2015,VQA}. The images in these datasets cover a broad range of categories and the questions are designed to test perceptual abilities such as counting, inferring spatial relationships, and identifying visual cues. Some  challenging questions require logical reasoning and memorization capabilities. However, the majority of the questions can be  answered by solely examining the visual content of the image. Hence, numerous approaches to solve these problems \cite{YuARXIV2015,VQA,goyal2017making,GaoNIPS2015,MalinowskiNIPS2014,MalinowskiICCV2015,hu2017learning} focus on extracting visual cues using deep networks. 

We note that many of the aforementioned methods focus on the visual aspect of the question answering task, \ie, the answer is predicted by combining representations of the question and the image. This clearly contrasts the described human-like approach, which combines observations with general knowledge. To address this discrepancy, in very recent meticulous work, Wang \etal~\cite{wang2018fvqa} introduced a `fact-based' VQA task (FVQA), an accompanying dataset, and a knowledge base of facts extracted from three different sources, namely WebChild~\cite{tandon2014webchild}, DBPedia~\cite{Auer2007dbpedia}, and ConceptNet~\cite{speer2017conceptnet}. Different from the classical VQA datasets, Wang \etal~\cite{wang2018fvqa} argued that such a dataset can be used to develop algorithms which answer more complex questions that require a combination of observation and general knowledge.
In addition to the dataset, Wang \etal~\cite{wang2018fvqa} also developed a model which leverages the information present in the supporting facts to answer questions about an image. 

To this end, Wang \etal~\cite{wang2018fvqa} design an approach which extracts keywords from the question and retrieves facts that contain those keywords from the knowledge base. Clearly, synonyms and homographs pose challenges which are hard to recover from.


\begin{figure}[t]
\centering
\includegraphics[width=\textwidth]{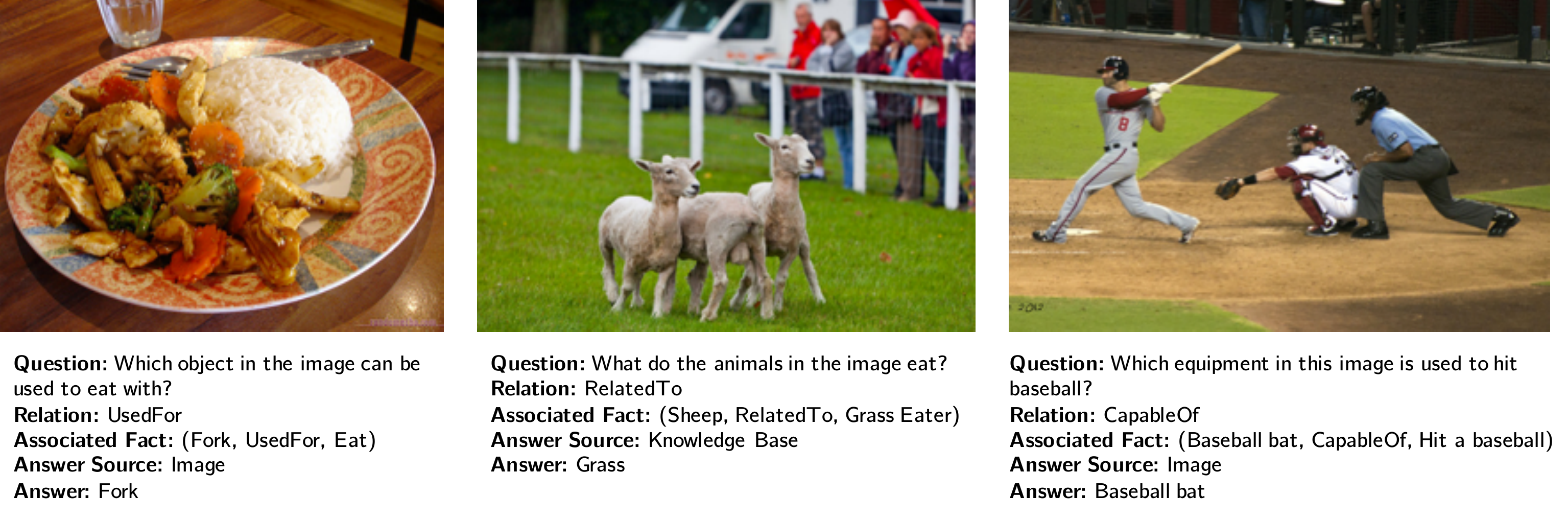}
\caption{The FVQA dataset expects methods to answer questions about images utilizing information from the image, as well as fact-based knowledge bases. Our method makes use of the image, and question text features, as well as high-level visual concepts extracted from the image in combination with a learned fact-ranking neural network. Our method is able to answer both visually grounded as well as fact based questions.}
\label{fig:teaser}
\end{figure}

To address this issue, we develop a learning based retrieval method. More specifically, our approach learns a parametric mapping of facts and question-image pairs to an embedding space. To answer a question, we use the fact that is most aligned with the provided question-image pair. As illustrated in \figref{fig:teaser}, our approach is able to accurately answer both more visual questions as well as more fact based questions. For instance, given the image illustrated on the left hand side along with the question, ``Which object in the image can be used to eat with?'', we are able to predict the correct answer, ``fork.'' Similarly, the proposed approach is able to predict the correct answer for the other two examples. 
Quantitatively we demonstrate the efficacy of the proposed approach on the recently introduced FVQA dataset, outperforming state-of-the-art by more than $5\%$ on the top-1 accuracy metric.

%% file: related.tex
\section{Related Work}
We develop a framework for visual question answering that benefits from a rich knowledge base. In the following, we first review classical visual question answering tasks before discussing visual question answering methods that take advantage of knowledge bases.

\medskip
\noindent{\bf Visual Question Answering.} 
In recent years, a significant amount of research has been devoted to developing techniques which can answer a question about a provided context such as an image. Of late, visual question answering has also been used to assess reasoning capabilities of state-of-the-art predictors. Using a variety of datasets~\cite{MalinowskiNIPS2014,RenNIPS2015,VQA,GaoNIPS2015,ZhuCVPR2016,JohnsonCVPR2017Clevr}, models based on multi-modal representation and attention~\cite{LuARXIV2016,YangCVPR2016,AndreasCVPR2016,DasARXIV2016,FukuiARXIV2016,ShihCVPR2016,XuARXIV2015,SchwartzNIPS2017}, deep network architectures~\cite{BenyounesICCV2017Mutan,MalinowskiICCV2015,MaARXIV2015,JainZhangCVPR2017}, and dynamic memory nets~\cite{XiongICML2016} have been developed. Despite these efforts, assessing the reasoning capabilities of present day deep network-based approaches and differentiating them from mere  memorization of training set statistics remains a hard task.
Most of the methods developed for visual question answering~\cite{RenNIPS2015,VQA,GaoNIPS2015,LuARXIV2016,YangCVPR2016,AndreasCVPR2016,DasARXIV2016,FukuiARXIV2016,ShihCVPR2016,XuARXIV2015,MalinowskiICCV2015,MaARXIV2015,XiongICML2016,KimARXIV2016,ZitnickAI2016,JabriARXIV2016,YuARXIV2015,ZhouARXIV2015,WuARXIV2016,JainCVPR2018} focus exclusively on answering questions related to observed content. To this end, these methods use image features extracted from networks such as the VGG-16~\cite{simonyan2014very} trained on large image datasets such as ImageNet~\cite{ILSVRC15}. However, it is unlikely that all the information which is required to answer a question is encoded in the features extracted from the image, or even the image itself. For example, consider an image containing a dog, and a question about this image, such as ``\emph{Is the animal in the image capable of jumping in the air}?''. In such a case, we would want our method to combine common sense and general knowledge about the world, such as the ability of a healthy dog to jump, along with features and observations from the image, such as the presence of the dog. This motivates us to develop methods that can use knowledge bases encoding general knowledge.

\medskip
\noindent{\bf Knowledge-based Visual Question Answering.} There has been interest in the natural language processing community in answering questions based on knowledge bases (KBs) using either semantic parsing~\cite{ZettlemoyerUAI2005,ZettlemoyerACL2005,BerantEMNLP2013,CaiACL2013,LiangCL2013,KwiatkowskiEMNLP2013,BerantACL2014,FaderKDD2014,YihACL2015,ReddyACL2016,XiaoACL2016} or information retrieval~\cite{UngerWWW2012,KolomiyetsIS2011,YaoACL2014,BordesEMNLP2014,BordesECML2014,DongACL2015,BordesICLR2015} methods. However, knowledge based visual question answering is still relatively unexplored, even though this is appealing from a practical standpoint as this decouples the reasoning by the neural network from the storage of knowledge in the KB. 
Notable examples in this direction are work by Zhu \etal~\cite{ZhuARXIV2015}, Wu \etal~\cite{WuCVPR2016}, Wang \etal~\cite{WangAI2017}, Krishnamurthy and Kollar~\cite{KrishnamurthyACL2013}, and Narasimhan \etal~\cite{NarasimhanEMNLP2016}. 


The works most related to our approach include \emph{Ask Me Anything} (AMA) by Wu \etal~\cite{wu2016ask}, Ahab by Wang \etal~\cite{Wang2017Ahab}, and FVQA by Wang \etal~\cite{wang2018fvqa}. AMA describes the content of an image in terms of a set of attributes predicted about the image, and  multiple captions generated about the image. The predicted attributes are used to query an external knowledge base, DBpedia~\cite{Auer2007dbpedia}, and the retrieved paragraphs are summarized to form a knowledge vector. The predicted attribute vector, the captions, and the database-based knowledge vector are passed as inputs to an LSTM that learns to predict the answer to the input question as a sequence of words. A drawback of this work is that it does not perform any explicit reasoning and ignores the possible structure in the KB. Ahab and FVQA, on the other hand, attempt to perform explicit reasoning. Ahab converts an input question into a database query, and processes the returned knowledge to form the final answer. Similarly, FVQA learns a mapping from questions to database queries through classifying questions into categories and extracting parts from the question deemed to be important. 
While both of these methods rely on fixed query templates, this very structure offers some insight into what information the method deems necessary to answer a question about a given image.
Both these methods use databases with a particular structure: those that contain facts about visual concepts represented as tuples, for example, (\emph{Cat}, \emph{CapableOf}, \emph{Climbing}), and (\emph{Dog}, \emph{IsA}, \emph{Pet}). We develop our method on the dataset released as part of the FVQA work, referred to as the FVQA dataset~\cite{wang2018fvqa}, which is a subset of three structured databases -- DBpedia~\cite{Auer2007dbpedia}, ConceptNet~\cite{speer2017conceptnet}, and WebChild~\cite{tandon2014webchild}.
The method presented in FVQA~\cite{wang2018fvqa} produces a query as an output of an LSTM which is fed the question as an input. 
Facts in the knowledge base are filtered on the basis of visual concepts such as objects, scenes, and actions extracted from the input image. The predicted query is then applied on the filtered database, resulting in a set of retrieved facts. A matching score is then computed between the retrieved facts and the question to determine the most relevant fact. The most correct fact forms the basis of the answer for the question.

In contrast to Ahab and FVQA, we propose to directly learn an embedding of facts and question-image pairs into a space that permits  to assess their compatibility. This has two important advantages over prior work: 1) by avoiding the generation of an explicit query, we eliminate errors due to synonyms, homographs, and incorrect prediction of visual concept type and answer type; and 2) our technique is easy to extend to any knowledge base, even one with a different structure or size. We also do not require any ad-hoc filtering of knowledge, and can instead learn to transform extracted visual concepts into a vector close to a relevant fact in the learned embedding space. Our method also naturally produces a ranking of facts deemed to be useful for the given question and image.

%% file: method.tex
\section{Learning Knowledge Base Retrieval}
\label{sec:method}
In the following, we first provide an overview of the proposed approach for knowledge based visual question answering before discussing our embedding space and learning formulation. 

\begin{figure}[t]
\centering
\includegraphics[trim={0, 0, 0cm, 0},width=\textwidth]{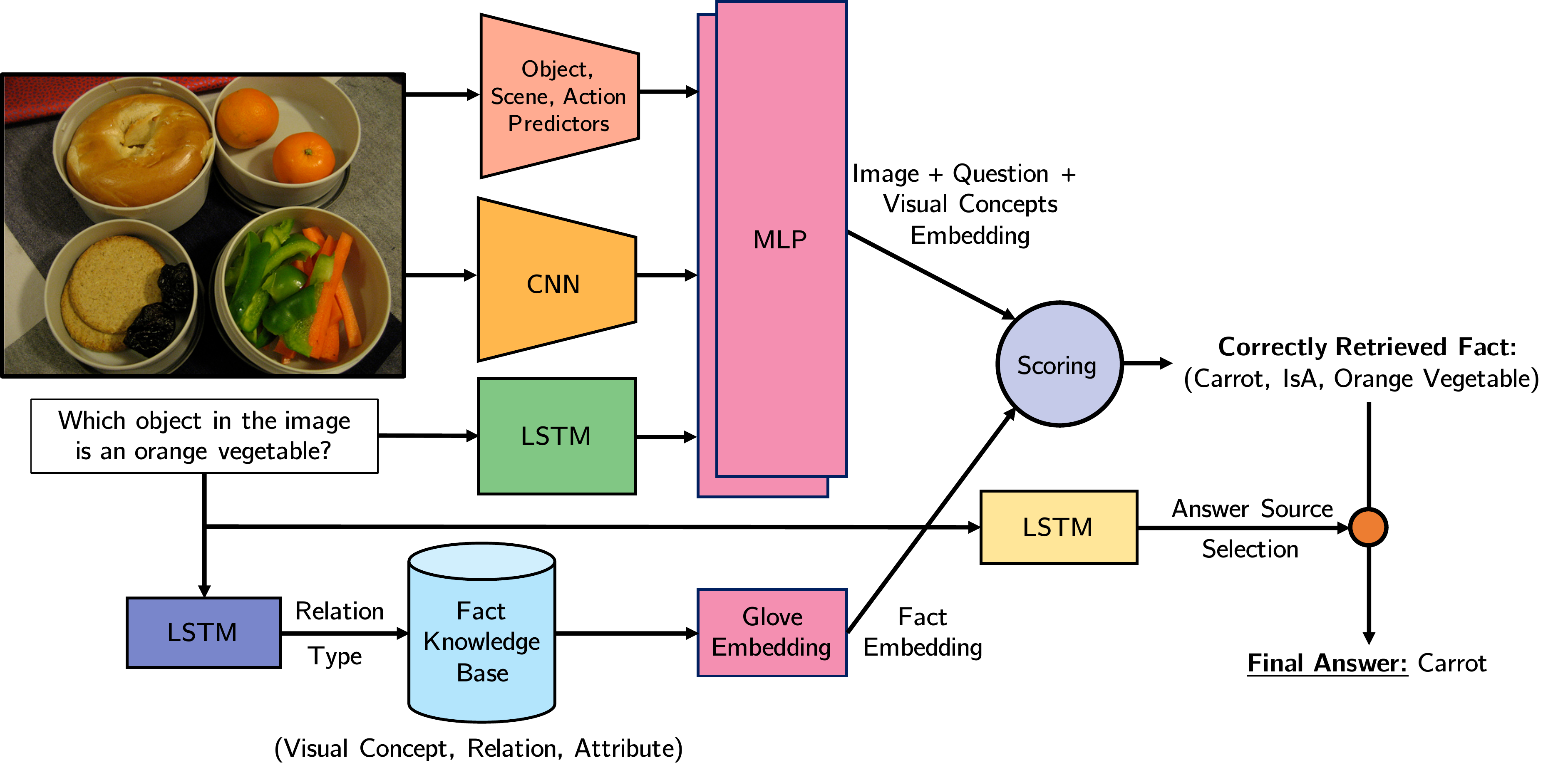}
\caption{Overview of the proposed approach. Given an image and a question about the image, we obtain an Image + Question Embedding through the use of a CNN on the image, an LSTM on the question, and a Multi Layer Perceptron (MLP) for combining the two modalities. In order to filter relevant facts from the Knowledge Base (KB), we use another LSTM to predict the fact relation type from the question. The retrieved structured facts are encoded using GloVe embeddings. The retrieved facts are ranked through a dot product between the embedding vectors and the top-ranked fact is returned to answer the question.}
\label{fig:overview}
\end{figure}

\noindent{\textbf{Overview.}} Our developed approach is outlined in~\figref{fig:overview}. The task at hand is to predict an answer $y$ for a question $Q$ given an image $x$ by using an external knowledge base KB, which consists of a set of facts $f_i$, \ie, $\text{KB} = \left\{ f_1, \ldots, f_{|\text{KB}|} \right\}$.
Each fact $f_i$ in the knowledge base is represented as a Resource Description Framework (RDF) triplet of the form $f_i = (a_i, r_i, b_i)$, where $a_i$ is a visual concept in the image, $b_i$ is an attribute or phrase associated with the visual entity $a_i$, and $r_i\in\cR$ is a relation between the two entities. 
The dataset contains $|\cR| = 13$  relations $r \in \cR = \{$\emph{Category}, \emph{Comparative}, \emph{HasA}, \emph{IsA}, \emph{HasProperty}, \emph{CapableOf}, \emph{Desires}, \emph{RelatedTo}, \emph{AtLocation}, \emph{PartOf}, \emph{ReceivesAction}, \emph{UsedFor}, \emph{CreatedBy}$\}$. 
Example triples of the knowledge base in our dataset are (\emph{Umbrella}, \emph{UsedFor}, \emph{Shade}), (\emph{Beach}, \emph{HasProperty}, \emph{Sandy}), (\emph{Elephant}, \emph{Comparative-LargerThan}, \emph{Ant}). 

To answer a question $Q$ correctly given an image $x$, we need to retrieve the right supporting fact and choose the correct entity, \ie, either $a$ or $b$. Importantly, entity $a$ is always derived from the image and entity $b$ is derived from the fact base. Consequently we refer to this choice as the answer source $s\in\left\{ \text{Image}, \text{KnowledgeBase} \right\}$. Using this formulation, we can extract the answer $y$ from a predicted fact $\hat f = (\hat a, \hat r, \hat b)$ and a predicted answer source $\hat s$ using
\begin{equation}
 y = 
 \begin{cases}
  \hat{a}, & \text{from } \hat{f} \text{ if } \hat{s} = \text{Image}\\
  \hat{b}, & \text{from } \hat{f} \text{ if } \hat{s} = \text{KnowledgeBase} 
 \end{cases}.
 \label{eq:anssource}
\end{equation}


It remains to answer, how to predict a fact $\hat f$ and how to infer the answer source $\hat s$. 
The latter is a binary prediction task and we describe our approach below. 
For the former, we note that the knowledge base contains a large number of facts. We therefore consider it infeasible to search through all the facts $f_i$ $\forall i\in \{1, \ldots, |\text{KB}|\}$ using an expensive evaluation based on a deep net. We therefore split this task into two parts: (1) Given a question, we train a network to predict the relation $\hat r$, that the question focuses on. 
(2) Using the predicted relation, $\hat r$, we reduce the fact space to those containing only the predicted relation. 

Subsequently, to answer the question $Q$ given image $x$,  we only assess the suitability of  the facts which contain the predicted relation $\hat r$. To assess the suitability, we design a score function $S(g^\text{F}(f_i), g^\text{NN}(x,Q))$ 
which measures the compatibility of a fact representation $g^\text{F}(f_i)$ and an image-question representation $g^\text{NN}(x,Q)$. Intuitively, the higher the score, the more suitable the fact $f_i$ for answering question $Q$ given image $x$.


Formally, we hence obtain the predicted fact $\hat f$ via
\be
\hat f = \arg\max_{i\in\{j : \operatorname{rel}(f_j) = \hat r\}} S(g^\text{F}(f_i), g^\text{NN}(x,Q)),
\label{eq:factinf}
\ee
where we search for the fact $\hat f$ maximizing the score $S$ among all facts $f_i$ which contain relation $\hat r$, \ie, among all $f_i$ with $i\in\{j : \operatorname{rel}(f_j) = \hat r\}$. Hereby we use the operator $\operatorname{rel}(f_i)$ to indicate the relation of the fact  triplet $f_i$. Given the predicted fact using \equref{eq:factinf} we obtain the answer $y$ from \equref{eq:anssource} after predicting the answer source $\hat s$.

This approach is outlined in \figref{fig:overview}. Pictorially, we illustrate the construction of an image-question embedding $g^\text{NN}(x,Q)$, via LSTM and CNN net representations that are combined via an MLP. We also illustrate the fact embedding $g^\text{F}(f_i)$. Both of them are combined using the score function $S(\cdot, \cdot)$, to predict a fact $\hat f$ from which we extract the answer as described in \equref{eq:anssource}.

In the following, we first provide details about the score function $S$, before discussing prediction of the relation $\hat r$ and prediction of the answer source $\hat s$.\

\noindent{\textbf{Scoring the facts.}} \figref{fig:overview} illustrates our approach to score the facts in the knowledge base, \ie, to compute $S(g^\text{F}(f_i), g^\text{NN}(x,Q))$. We obtain the score in three steps: (1) computing of a fact representation $g^\text{F}(f_i)$; (2) computing of an image-question representation $g^\text{NN}(x,Q)$; (3) combination of the fact and image-question representation to obtain the final score $S$. We discuss each of those steps in the following.

\noindent\emph{(1) Computing a fact representation.} 
To obtain the fact representation $g^\text{F}(f_i)$, we concatenate two vectors, the averaged GloVe-100~\cite{pennington2014glove} representation of the words of entity $a_i$ and the averaged GloVe-100 representation of the words of entity $b_i$. Note that this fact representation is non-parametric, \ie, there are no trainable parameters.

\noindent\emph{(2) Computing an image-question representation.} 
We compute the image-question representation $g^\text{NN}(x,Q)$, by combining a visual representation $g_w^V(x)$, obtained from a standard deep net, \eg, ResNet or VGG, with a visual concept representation $g_w^C(x)$, and a sentence representation $g_w^Q(Q)$, of the question $Q$, obtained using a trainable recurrent net. For notational convenience we concatenate all trainable parameters into one vector $w$. Making the dependence on the parameters explicit, we obtain the image-question representation via
$g^\text{NN}_w(x, Q) = g^\text{NN}_w(g_w^V(x), g_w^Q(Q), g_w^C(x)).$

More specifically, for the question embedding $g^Q_w(Q)$, we use an LSTM model~\cite{HochreiterNC1997}. 
For the image embedding $g^V_w(x)$, we extract image features using ResNet-152~\cite{he2016deep} pre-trained on the ImageNet dataset~\cite{deng2009imagenet}. 
In addition, we also extract a visual concept representation $g_w^C(x)$, which is a multi-hot vector of size 1176 indicating the visual concepts which are grounded in the image. The visual concepts detected in the images are objects, scenes, and actions. For \emph{objects}, we use the detections from two Faster-RCNN~\cite{NIPS2015_5638} models that are trained on the Microsoft COCO 80-object~\cite{coco} and the ImageNet 200-object~\cite{ILSVRC15} datasets. In total, there are 234 distinct object classes, from which we use that subset of labels that coincides with the FVQA dataset. The \emph{scene} information (such as pasture, beach, bedroom) is extracted by the VGG-16 model~\cite{simonyan2014very} trained on the MIT Places 365-class dataset~\cite{zhou2017places}. Again, we use a subset of Places to construct the 1176-dimensional multi-hot vector $g_w^C(x)$. For detecting \emph{actions}, we use the CNN model proposed in~\cite{mallya2016learning} which is trained on the HICO~\cite{chao:iccv2015} and MPII~\cite{andriluka14cvpr} datasets. The HICO dataset contains labels for 600 human-object interaction activities while the MPII dataset contains labels for 393 actions. We use a subset  of actions, namely those which coincide with the ones in the FVQA dataset. 


All the three vectors $g_w^V(x), g_w^Q(Q), g_w^C(x)$ are concatenated and passed to the multi-layer perceptron $g^\text{NN}_w(\cdot, \cdot, \cdot)$. 

\noindent\emph{(3) Combination of fact and image-question representation.}
For each fact representation $g^\text{F}(f_i)$, we compute a score
$$
S_w(g^\text{F}(f_i), g_w^\text{NN}(x,Q)) = \cos(g^\text{F}(f_i), g_w^\text{NN}(x,Q)) = \frac {g^\text{F}(f_i) \cdot g_w^\text{NN}(x,Q)}{||g^\text{F}(f_i)|| \cdot ||g_w^\text{NN}(x,Q)||},
$$
where $g_w^\text{NN}(x,Q)$ is the image question representation. Hence, the score $S$ is the cosine similarity between the two normalized representations and represents the fit of fact $f_i$ to the image-question pair $(x,Q)$. 

\noindent{\textbf{Predicting the relation.}} To predict the relation $\hat r\in\cR = h_{w_1}^r(Q)$, from the obtained question $Q$, we use an LSTM net. More specifically, we first embed and then encode the words of the question $Q$, one at a time, and linearly transform the final hidden representation of the LSTM to predict $\hat r$, from $|\cR|$ possibilities using a standard multinomial classification. For the results presented in this work, we trained the relation prediction parameters $w_1$ independently of the score function. We leave a joint formulation to future work. 

\noindent{\textbf{Predicting the answer source.}} Prediction of the answer source $\hat s = h_{w_2}^s(Q)$ from a given question $Q$ is similar to relation prediction. Again, we use an LSTM net to embed and encode the words of the question $Q$ before linearly transforming the final hidden representation to predict $\hat s\in\{\text{Image}, \text{KnowledgeBase}\}$. Analogous to relation prediction, we train this LSTM net's parameters $w_2$ separately and leave a joint formulation to future work.

\begin{algorithm}[t]
\caption{Training with hard negative mining}\label{alg:ours}
\hspace*{\algorithmicindent} \textbf{Input:} $(x, Q, f^\ast)$, $KB$\\
\hspace*{\algorithmicindent} \textbf{Output:} parameters $w$ 
\begin{algorithmic}[1]
\FOR{$t = 0, \ldots, T$}
	\STATE Create dataset $\cD^{(t)}$ by sampling negative facts randomly (if $t = 0$) or by retrieving facts predicted wrongly with $w^{(t-1)}$ (if $t > 0$)
	\STATE Use $\cD^{(t)}$ to obtain $w^{(t)}$ by optimizing the program given in \equref{eq:finallearn}
\ENDFOR
\RETURN $w^{(T)}$
\end{algorithmic}
\end{algorithm}

\noindent{\textbf{Learning.}} As mentioned before, we train the parameters $w$ (score function), $w_1$ (relation prediction), and $w_2$ (answer source prediction) separately. To train $w_1$, we use a dataset $\cD_1 = \{(Q, r)\}$ containing pairs of question and the corresponding relation which was used to obtain the answer. To learn $w_2$, we use a dataset $\cD_2 = \{(Q, s)\}$, containing pairs of question and the corresponding answer source. For both classifiers we use stochastic gradient descent on the classical cross-entropy and binary cross-entropy loss respectively. Note that both the datasets are readily available from~\cite{wang2018fvqa}.

To train the parameters of the score function we adopt a successive approach operating in time steps $t = \{1, \ldots, T\}$. In each time step, we gradually increase the difficulty of the dataset $\cD^{(t)}$ by mining hard negatives. More specifically, for every question $Q$, and image $x$, $\cD^{(0)}$ contains the `groundtruth' fact $f^\ast$ as well as 99 randomly sampled `non-groundtruth' facts. After having trained the score function on this dataset we use it to predict facts for image-question pairs and create a new dataset $\cD^{(1)}$ which now contains, along with the groundtruth fact, another 99 non-groundtruth facts that the score function assigned a high score to. 

Given a dataset $\cD^{(t)}$, we train the parameters $w$ of the representations involved in the score function $S_w(g^\text{F}(f_i), g^\text{NN}_w(x,Q))$, and its image, question, and concept embeddings by encouraging that the score of the groundtruth fact $f^\ast$ is larger than the score of any other fact. More formally, we aim for parameters $w$ which ensure the classical margin, \ie, an SVM-like loss for deep nets:
\be
S_w(f^\ast, x, Q) \geq L(f^\ast, f) + S_w(f, x, Q) \quad\quad \forall (f, x, Q) \in \cD^{(t)},
\label{eq:margin}
\ee
where $L(f^\ast, f)$ is the task loss (aka margin) comparing the groundtruth fact $f^\ast$ to other facts $f$. In our case $L\equiv 1$. Since we may not find parameters $w$ which ensure feasibility $\forall (f, x, Q) \in \cD^{(t)}$, we introduce slack variables $\xi_{(f,x,Q)} \geq 0$ to obtain after reformulation:
\be
\xi_{(f,x,Q)} \geq L(f^\ast, f) + S_w(f, x, Q) - S_w(f^\ast, x, Q) \quad\quad \forall (f, x, Q) \in \cD^{(t)}.
\label{eq:margin1}
\ee
Instead of enforcing the constraint $\forall (f, x, Q)$ in the dataset $\cD^{(t)}$, it is equivalent to require~\cite{Tsochantaridis2005}
\be
\xi_{(x,Q)} \geq \max_f \{L(f^\ast, f) + S_w(f, x, Q)\} - S_w(f^\ast, x, Q) \quad\quad \forall (x, Q) \in \cD^{(t)}.
\label{eq:margin2}
\ee
Using this constraint, we find the parameters $w$ by solving
\be
\min_{w, \xi_{(x,Q)}\geq 0} \frac{C}{2}\|w\|_2^2 + \sum_{(x, Q)\in\cD^{(t)}} \xi_{(x,Q)} \quad\text{s.t. Constraints in \equref{eq:margin2}.}
\label{eq:prog1}
\ee
For applicability of the standard sub-gradient descent techniques, we reformulate the program given in \equref{eq:prog1} to read as
\be
\min_{w} \frac{C}{2}\|w\|_2^2 + \sum_{(x, Q)\in\cD^{(t)}} \left(\max_f \{L(f^\ast, f) + S_w(f, x, Q)\} - S_w(f^\ast, x, Q)\right),
\label{eq:finallearn}
\ee
which can be optimized using standard deep net packages. The proposed approach for learning the parameters $w$ is summarized in \algref{alg:ours}. 
In the following we now assess the suitability of the proposed approach.

%
%
%
%

%% file: evaluation.tex
\section{Evaluation}
In the following, we assess the proposed approach. We first provide details about the proposed dataset before presenting quantitative results for prediction of relations from questions, prediction of answer-source from questions, and prediction of the answer and the supporting fact. We also discuss mining of hard negatives. Finally, we show qualitative results. 

\noindent{\textbf{Dataset and Knowledge Base.}} We use the publicly available FVQA dataset~\cite{wang2018fvqa} and its knowledge base to evaluate our model. This dataset consists of 2,190 images, 5,286 questions, and 4,126 unique facts corresponding to the questions. The knowledge base, consisting of 193,449 facts,  were constructed by extracting the top visual concepts for all the images in the dataset and querying for those concepts in the three knowledge bases, WebChild~\cite{tandon2014webchild}, ConceptNet~\cite{speer2017conceptnet}, and DBPedia~\cite{Auer2007dbpedia}.  The dataset consists of 5 train-test folds, and all the scores we report are averaged across all splits.

\noindent{\textbf{Predicting Relations from Questions.}} We use an LSTM architecture as discussed in \secref{sec:method} 
to predict the relation $r\in\cR$ given a question $Q$. The standard train-test split of the FVQA dataset is used to evaluate our model. Batch gradient descent with Adam optimizer was used on batches of size 100 and the model was trained over 50 epochs. LSTM embedding and word embeddings are of size 128 each. The learning rate is set to $1\mathrm{e}{-3}$ and a dropout of 0.7 is applied after the word embeddings as well as the LSTM embedding. Table~\ref{table:relation_results} provides a comparison of our model to the  FVQA baseline~\cite{wang2018fvqa} using top-1 and top-3 prediction accuracy. We observe our results to improve the baseline by more than 10\% on top-1 accuracy and by more than 9\% when using the top-3 accuracy metric. 

\begin{table}[t]
\centering
\begin{tabular}{cc}
\begin{minipage}[b]{0.50\linewidth}
{
\centering
 \setlength{\tabcolsep}{10pt}
 \begin{tabular}{l|c|c}\toprule
   \multirow{2}{*}{\bf Method} & \multicolumn{2}{c}{\bf Accuracy} \\
   & {\bf @1} & {\bf @3} \\\midrule
   FVQA~\cite{wang2018fvqa} & 64.94 & 82.42 \\\hline
   Ours & {\bf 75.4} & {\bf 91.97} \\\bottomrule
 \end{tabular}
 \caption{Accuracy of predicting relations given the question.}
 \label{table:relation_results}
 } 
 \end{minipage}
 &
 \begin{minipage}[b]{0.50\linewidth}
 {
 \centering
  \setlength{\tabcolsep}{10pt}
 \begin{tabular}{l|c|c}\toprule
   \multirow{2}{*}{\bf Method} & \multicolumn{2}{c}{\bf Accuracy} \\
   & {\bf @1} & {\bf @3} \\\midrule
   Ours & 97.3 & 100.00 \\\bottomrule
 \end{tabular}
 \caption{Accuracy of predicting answer source from a given question.}
 \label{table:ans_results}
 }
 \end{minipage}
\end{tabular}
\end{table}
\noindent{\textbf{Predicting Answer Source from Questions.}} We assess the accuracy of predicting the answer source $s$ given a question $Q$. To predict the source of the answer, we use an LSTM architecture as discussed in detail in \secref{sec:method}. 
Note that for predicting the answer source, the size of the LSTM embedding and word embeddings was set to 64 each. Table~\ref{table:ans_results} summarizes the accuracy of the prediction results 
 of our model. We observe the prediction accuracy of the proposed approach to be close to perfect.

\noindent{\textbf{Predicting the Correct Answer.}} 
Our score function based model to retrieve the supporting fact is described in detail in \secref{sec:method}. 
For the image embedding, we pass the 2048 dimensional feature vector returned by ResNet through a fully-connected layer and reduce it to a 64 dimensional vector. For the question embedding, we use an LSTM with a hidden layer of size 128. The two are then concatenated into a vector of size 192 and passed through a two layer perceptron with 256 and 128 nodes respectively. Note that the baseline doesn't use image features apart from the detected visual concepts.

The multi-hot visual concept embedding is passed through a fully-connected layer to form a 128 dimensional vector. This is then concatenated with the output of the perceptron and passed through another layer with 200 output nodes. We found a late fusion of the visual concepts to results in a better model as the facts explicitly contain these terms.

Fact embeddings are constructed using GloVe-100 vectors each, for entities $a$ and $b$. If $a$ or $b$ contain multiple words, an average of all the embeddings is computed. We use cosine distance between the MLP and the fact embeddings to score the facts. The highest scoring fact is chosen as the answer. Ties are broken randomly. 

Based on the answer source prediction which is computed using the aforementioned LSTM model, we choose either entity $a$ or $b$ of the fact to be the answer. See \equref{eq:anssource} for the formal description. Accuracy is computed based on exact match between the chosen entity and the groundtruth answer. 

To assess the importance of particular features we investigate 5 variants of our model with varying features: two oracle approaches `$gt$ Question + Image + Visual Concepts' and `$gt$ Question + Visual Concepts' which make use of groundtruth relation type and answer type data. More specifically, `$gt$ Question + Image + Visual Concepts' and `$gt$ Question + Visual Concepts' use the groundtruth relations and answer sources respectively. We have three approaches using a variety of features as follows: `Question + Image + Visual Concepts,' `Question + Visual Concepts,' and `Question + Image.'
We drop either the Image embeddings from ResNet or the Visual Concept embeddings to obtain two other models, `Question + Visual Concepts' and `Question + Image.'

Table~\ref{table:answer_results} shows the accuracy of our model in predicting an answer and compares our results to other FVQA baselines. We observe the proposed approach to outperform the state-of-the-art ensemble technique by more than $3\%$ and the strongest baseline without ensemble by over $5\%$ on the top-1 accuracy metric. Moreover we note the importance of visual concepts to accurately predict the answer. By including groundtruth information we assess the maximally possible top-1 and top-3 accuracy. We observe the difference to be around $8\%$, suggesting that there is some room for improvement. 
 
\noindent{\textbf{Question to Supporting Fact.}} To provide a complete assessment of the proposed approach we illustrate in 
Table~\ref{table:retrieval_results} the top-1 and top-3 accuracy scores in retrieving the supporting facts of our model compared to other FVQA baselines. We observe the proposed approach to improve significantly both the top-1 and top-3 accuracy by more than $20\%$. We think this is a significant improvement towards efficiently including knowledge bases into visual question answering. 
\begin{table}[t]
 \centering
  \setlength{\tabcolsep}{10pt}
 \begin{tabular}{l|c|c}\toprule
   \multirow{2}{*}{\bf Method} & \multicolumn{2}{c}{\bf Accuracy} \\
   & {\bf @1} & {\bf @3} \\\midrule
   LSTM-Question+Image+Pre-VQA~\cite{wang2018fvqa} & 24.98 & 40.40 \\\hline
   Hie-Question+Image+Pre-VQA~\cite{wang2018fvqa} & 43.14 & 59.44 \\\hline
   FVQA~\cite{wang2018fvqa} & 56.91 & 64.65 \\\hline
   Ensemble~\cite{wang2018fvqa} & 58.76 & - \\\midrule\midrule
   Ours - Question + Image & 26.68 & 30.27 \\\hline
   Ours - Question + Image + Visual Concepts & 60.30 & 73.10 \\\hline
   Ours - Question + Visual Concepts & {\bf 62.20} & {\bf 75.60} \\\midrule\midrule
   Ours - $gt$ Question + Image + Visual Concepts & 69.12 & 80.25 \\\hline
   Ours - $gt$ Question + Visual Concepts & 70.34 & 82.12 \\\bottomrule
 \end{tabular}
 \caption{Answer accuracy over the FVQA dataset.}
 \label{table:answer_results}
 \centering
  \setlength{\tabcolsep}{10pt}
 \begin{tabular}{l|c|c}\toprule
   \multirow{2}{*}{\bf Method} & \multicolumn{2}{c}{\bf Accuracy} \\
   & {\bf @1} & {\bf @3} \\\midrule
   FVQA-top-1~\cite{wang2018fvqa} & 38.76 &42.96 \\\hline
   FVQA-top-3~\cite{wang2018fvqa} & 41.12 & 45.49 \\\midrule\midrule
   Ours - Question + Image & 28.98 & 32.34 \\\hline
   Ours - Question + Image + Visual Concepts & 62.30 & 74.90 \\\hline
   Ours - Question + Visual Concepts & {\bf 64.50} & {\bf 76.20} \\\bottomrule
 \end{tabular}
 \caption{Correct fact prediction precision over the FVQA dataset.}
 \label{table:retrieval_results}
 \centering
  \setlength{\tabcolsep}{10pt}
 \begin{tabular}{l|c|c|c}\toprule
   \multirow{2}{*}{\bf Iteration} &\multirow{2}{*}{\bf \# Hard Negatives} &\multicolumn{2}{c}{\bf Precision} \\
   & & {\bf @1} & {\bf @3} \\\midrule
   1 & 0 & 20.17 & 23.46 \\\hline
   2 & 84,563 & 38.65 & 45.49 \\\hline
   3 & 6,889 & 64.5 & 76.2 \\\bottomrule
 \end{tabular}
 \caption{Correct fact prediction precision with hard negative mining.}
 \label{table:hard_negs}
\end{table}
\begin{figure}[t]
 \centering
 \includegraphics[width=0.95\textwidth]{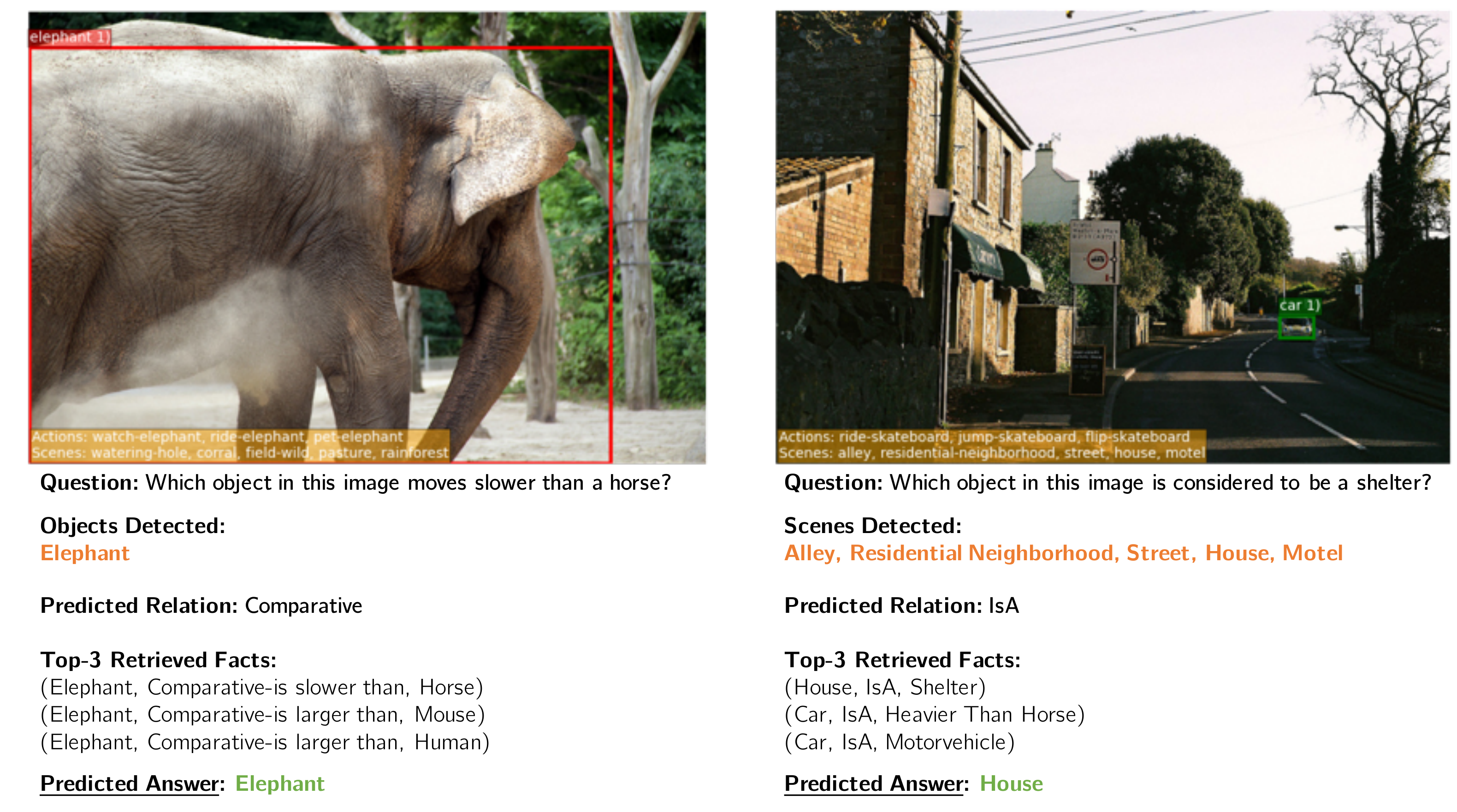}
 \caption{Examples of Visual Concepts (VCs) detected by our framework. Here, we show examples of detected objects, scenes, and actions predicted by the various networks used in our pipeline. There is a clear alignment between useful facts, and the predicted VCs. As a result, including VCs in our scoring method helps improve performance.}
 \label{fig:vc_predictions}
\end{figure}
\begin{figure}[t]
 \centering
 \includegraphics[width=0.95\textwidth]{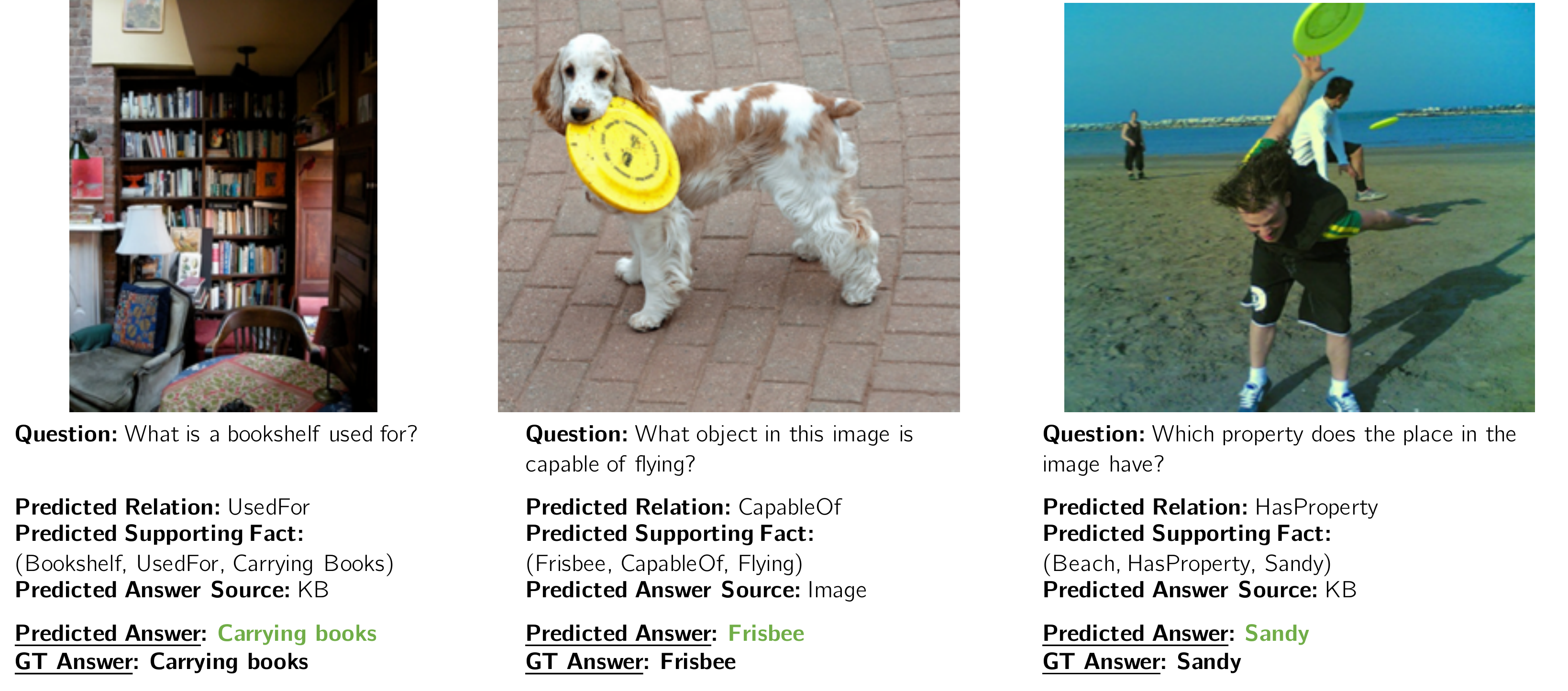}\\
 \includegraphics[width=0.95\textwidth]{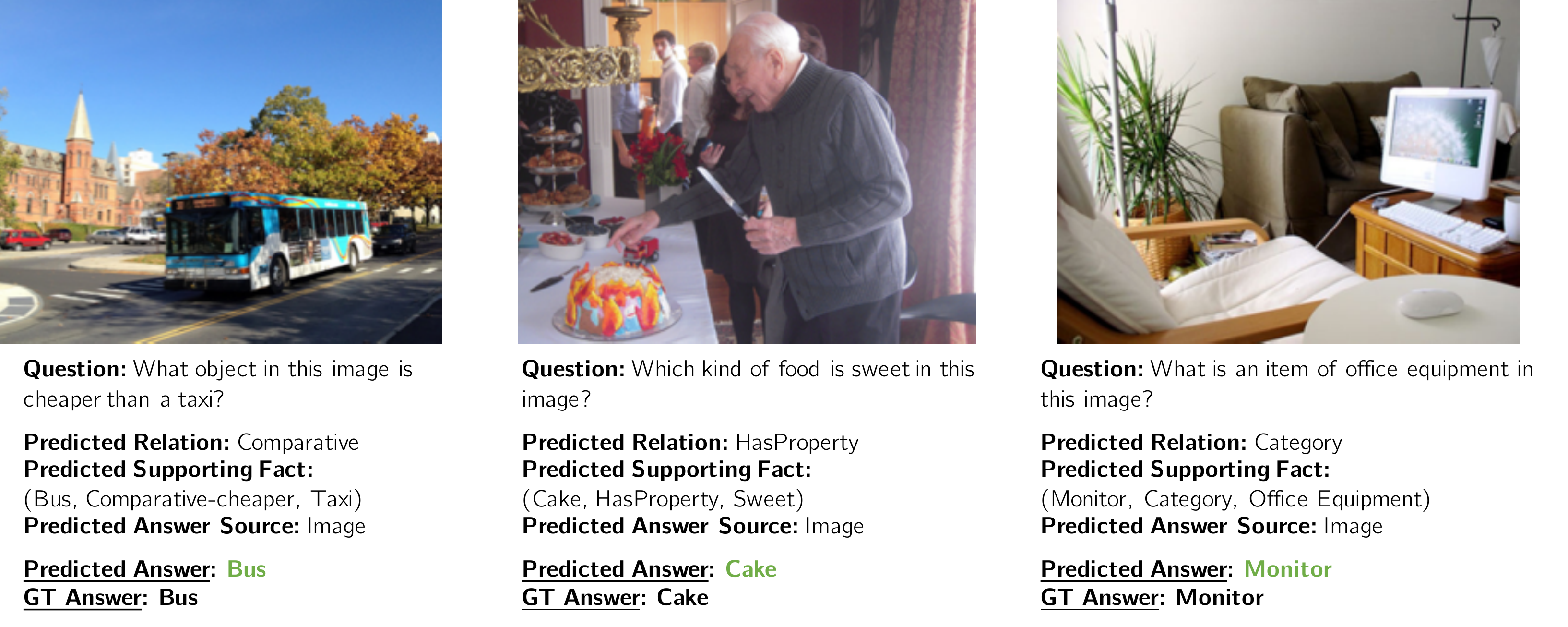}\\
 \rule{\textwidth}{1pt}\\[5pt]
 
 \includegraphics[width=0.95\textwidth]{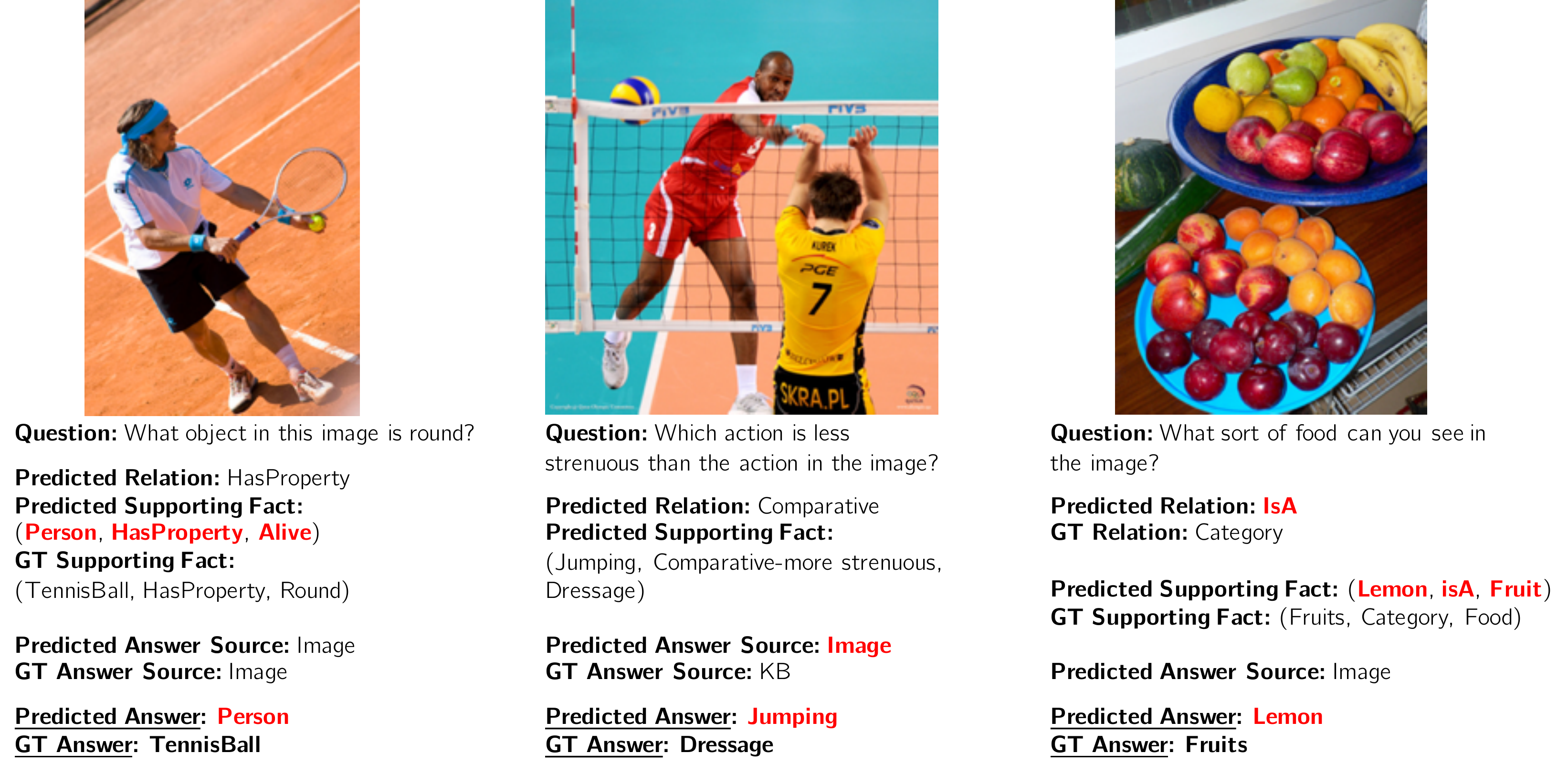}
 \caption{Success and failure cases of our method. In the top two rows, our method correctly predicts the relation, the supporting fact, and the answer source to produce the correct answer for the given question. The bottom row of examples shows the failure modes of our method.
 }
 \label{fig:qualitative_examples}
\end{figure}

\noindent{\textbf{Mining Hard Negatives.}} We trained our model over three iterations of hard negative mining, \ie, $T = 2$. In iteration 1 ($t = 0$), all the 193,449 facts were used to sample the 99 negative facts during train. At every 10th epoch of training, negative facts which received high scores were saved. In the next iteration, the trained model along with the negative facts is loaded and we ensure that the 99 negative facts are now sampled from the hard negatives. Table~\ref{table:hard_negs} shows the Top-1 and Top-3 accuracy for predicting the supporting facts over each of the three iterations. We observe significant improvements due to the proposed hard negative mining strategy. While na\"ive training of the proposed approach yields  only $20.17\%$ top-1 accuracy, two iterations improve the performance to $64.5\%$. 

\noindent{\textbf{Synonyms and Homographs.}} Here we show the improvements of our model compared to the baseline with respect to synonyms and homographs. To this end, we run additional tests using Wordnet to determine the number of question-fact pairs which contain synonyms. The test data contains 1105 such pairs out of which our model predicts 91.6\% (1012) correctly, whereas the FVQA model predicts 78.0\% (862) correctly. In addition, we manually generated 100 synonymous questions by replacing words in the questions with synonyms (\eg, ``What in the bowl can you eat?" is rephrased to ``What in the bowl is edible?"). Tests on these 100 new samples find that our model predicts 89 of these correctly, whereas the key-word matching FVQA technique~\cite{wang2018fvqa} gets 61 of these right.
With regards to homographs, the test set has 998 questions which contain words that have multiple meanings across facts. Our model predicts correct answers for  79.4\% (792), whereas the FVQA model gets  66.3\% (662) correct.

\noindent{\textbf{Qualitative Results.}}
\figref{fig:vc_predictions} shows the Visual Concepts (VCs) detected for a few samples along with the top 3 facts retrieved by our model. Providing these predicted VCs as input to our fact-scoring MLP helps improve supporting fact retrieval as well as answer accuracy by a large margin of over $30\%$ as seen in Tables~\ref{table:answer_results} and~\ref{table:retrieval_results}. As can be seen in~\figref{fig:vc_predictions}, there is a close alignment between relevant facts and predicted VCs, as VCs provide a high-level overview of the salient content in the images.

In~\figref{fig:qualitative_examples}, we show success and failure cases of our method. There are 3 steps to producing the correct answer using our method: (1) correctly predicting the relation, (2) retrieving supporting facts containing the predicted relation, and relevant to the image, and (3) choosing the answer from the predicted answer source (Image/Knowledge Base). The top two rows of images show cases where all the 3 steps were correctly executed by our proposed method. Note that our method works for a variety of relations, objects, answer sources, and varying difficulty. It is correctly able to identify the object of interest, even when it is not the most prominent object in the image. For example, in the middle image of the first row, the frisbee is smaller than the dog in the image. However, we were correctly able to retrieve the supporting fact about the frisbee using information from the question, such as `\emph{capable of}' and `\emph{flying}.'

A mistake in any of the 3 steps can cause our method to produce an incorrect answer. The bottom row of images in~\figref{fig:qualitative_examples} displays prototypical failure modes. In the leftmost image, we miss cues from the question such as `\emph{round},' and instead retrieve a fact about the person. In the middle image, our method makes a mistake at the final step and uses information from the wrong answer source. This is a very rare source of errors overall, as we are over $97\%$ accurate in predicting the answer source, as shown in Table~\ref{table:ans_results}.
In the rightmost image, our method makes a mistake at the first step of predicting the relation, making the remaining steps incorrect. Our relation prediction is around $75\%$, and $92\%$ accurate by the top-1 and top-3 metrics, as shown in Table~\ref{table:relation_results}, and has some scope for improvement. For qualitative results regarding synonyms and homographs we refer the interested reader to the supplementary material. 



%% file: conclusion.tex
\section{Conclusion}
In this work, we  addressed knowledge-based visual question answering and developed a method that learns to embed facts as well as question-image pairs into a space that admits efficient search for answers to a given question. In contrast to existing retrieval based techniques, our approach learns to embed questions and facts for retrieval. We have demonstrated the efficacy of the proposed method on the recently introduced and challenging FVQA dataset, producing state-of-the-art results. 
In the future, we hope to address extensions of our work to larger structured knowledge bases, as well as unstructured knowledge sources, such as online text corpora.

\noindent\textbf{Acknowledgments:} This material is based upon work supported in part by the National Science Foundation under Grant No.~1718221, Samsung, and 3M. We thank NVIDIA for providing the GPUs used for this research. We also thank Arun Mallya and Aditya Deshpande for their help.